\pdfoutput=1
\documentclass[11pt]{article}

\RequirePackage[dvipsnames,table]{xcolor}

\usepackage[]{acl}

\usepackage{times}
\usepackage{latexsym}

\usepackage[T1]{fontenc}
\usepackage[utf8]{inputenc}

\usepackage{microtype}

\usepackage{inconsolata}

\usepackage{graphicx}
\usepackage{amsmath}
\usepackage{amssymb}
\usepackage{booktabs}
\usepackage{multirow} 
\usepackage{tikz}               % for drawing the panels
\usetikzlibrary{calc}           % for coordinate calculations
\usepackage{enumitem}           % to customize the enumerate in the plan
\usetikzlibrary{positioning}    % for the “below=… of …” and “right=… of …” keys
\usepackage[most]{tcolorbox}
\usepackage{geometry}
\usepackage{subcaption}

\definecolor{darkgolden}{HTML}{B8860B}

\title{\textsc{Plan-Tuning}: Post-Training Language Models to Learn Step-by-Step Planning for Complex Problem Solving}

\author{Mihir Parmar$^{1}$ \quad Palash Goyal$^1$ \quad Xin Liu$^1$ \quad Yiwen Song$^1$ \quad Mingyang Ling$^1$\\\textbf{Chitta Baral}$^2$ \quad \textbf{Hamid Palangi}$^{1*}$ \quad \textbf{Tomas Pfister}$^{1*}$ \\\\
$^1$Google \quad $^2$Arizona State University
}

\begin{document}
\maketitle
\begin{abstract}

Recently, decomposing complex problems into simple subtasks--a crucial part of human-like natural planning--to solve the given problem has significantly boosted the performance of large language models (LLMs). However, leveraging such planning structures during post-training to boost the performance of smaller open-source LLMs remains underexplored. Motivated by this, we introduce \textsc{Plan-Tuning}, a unified post-training framework that (i) distills synthetic task decompositions (termed “planning trajectories”) from large-scale LLMs and (ii) fine-tunes smaller models via supervised and reinforcement-learning objectives designed to mimic these planning processes to improve complex reasoning. On GSM8k and the MATH benchmarks, plan-tuned models outperform strong baselines by an average $\sim7\%$. Furthermore, plan-tuned models show better generalization capabilities on out-of-domain datasets, with average $\sim10\%$ and $\sim12\%$ performance improvements on OlympiadBench and AIME 2024, respectively. Our detailed analysis demonstrates how planning trajectories improves complex reasoning capabilities, showing that \textsc{Plan-Tuning} is an effective strategy for improving task-specific performance of smaller LLMs.

% \makeatletter
% % ——— First special footnote with “*” ———
% \begingroup
%   \renewcommand\thefootnote{*}%
%   \footnotetext{%
%     Please correspond with Mihir Parmar%
%     \href{mailto:mihirparmar@asu.edu}{<mihirparmar@asu.edu>},%
%     and Hamid Palangi%
%     \href{mailto:hamidpalangi@google.com}{<hamidpalangi@google.com>}%
%   }%
% \endgroup
% \setcounter{footnote}{0}% reset for the next one

% % ——— Second special footnote with “♣” ———
% \begingroup
%   \renewcommand\thefootnote{$\clubsuit$}%
%   \footnotetext{Joint last authors}%
% \endgroup
% \setcounter{footnote}{0}% back to normal numbering

% % After these groups, \thefootnote and the counter are exactly as before.
% \makeatother

\def\thefootnote{$\dagger$}\footnotetext{Please correspond with \texttt{\{mihirparmar,hamidpalangi\}@google.com}}\def\thefootnote{\english{footnote}}
\def\thefootnote{$*$}\footnotetext{Joint last authors}\def\thefootnote{\english{footnote}}

\end{abstract}

\section{Introduction}
\label{sec:intro}

\begin{figure}
    \centering
    \includegraphics[width=\linewidth]{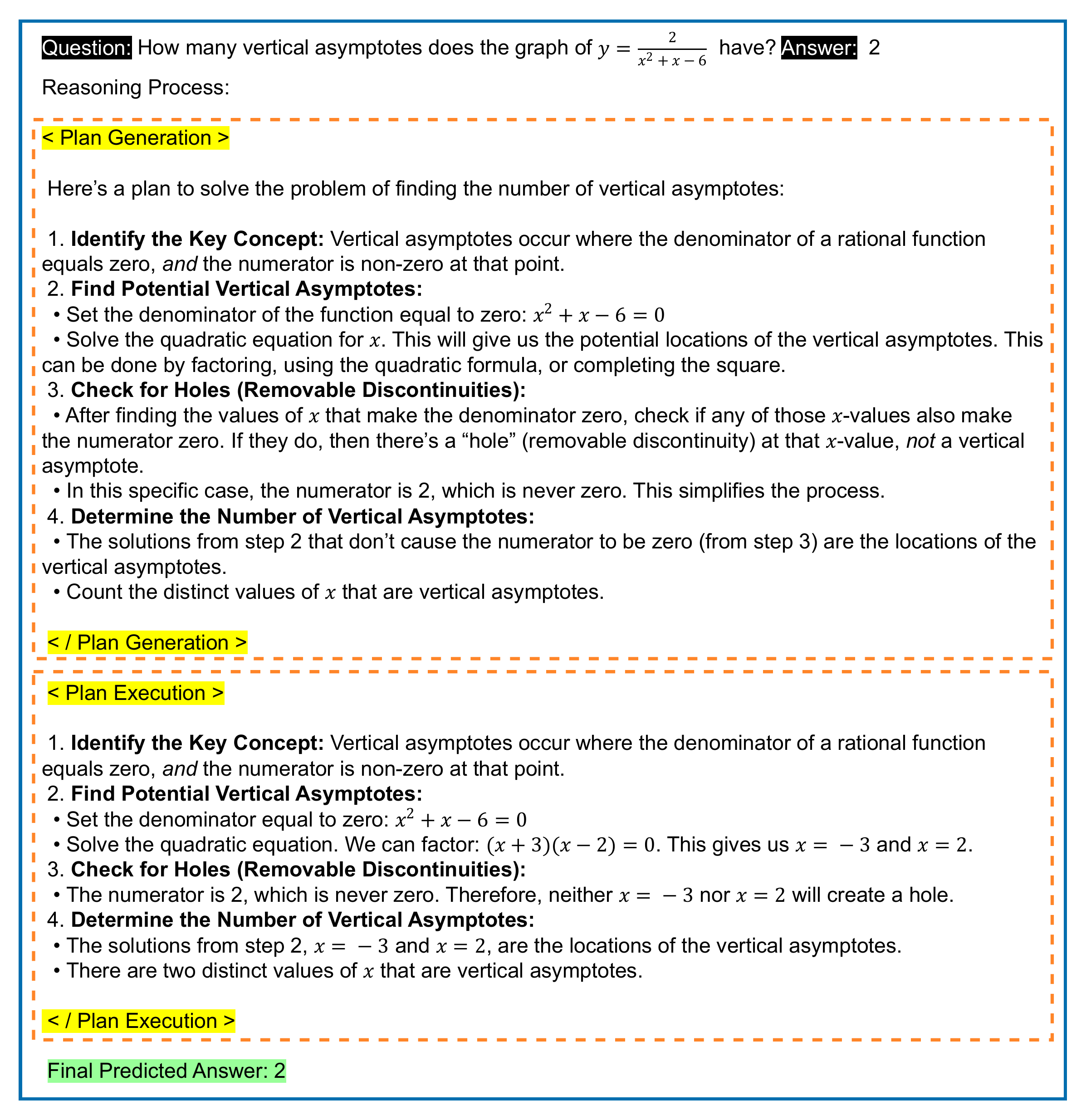}
    \caption{An example planning trajectory for a problem adapted from the MATH \cite{hendrycks2021measuring}.}
    \label{fig:example}
\end{figure}

Natural planning aligns more with real-world tasks such as trip or meeting planning \cite{zheng2024natural}. Decomposing complex problems into simpler subtasks is a key to human-like natural planning \cite{jiao-etal-2024-learning, parmar2025plangen}. For instance, when prompted with \textit{“How to plan a trip to …”}, recent LLMs naturally generate subtasks such as selecting a mode of transport, estimating a budget, and determining the trip duration (see App. \ref{app:examples_planning} for examples). Recent large-scale LLMs such as o4 \cite{zhong2024evaluation}, Gemini-2.0-Pro \cite{team2023gemini}, and Deepseek-v3 \cite{liu2024deepseek} have demonstrated this ability, and it significantly boosts their performance on complex reasoning tasks \cite{rein2024gpqa, he2024olympiadbench}. Figure \ref{fig:example} illustrates a synthetic planning trajectory distilled from Gemini-2.0-Pro: a complex problem is first decomposed into intermediate subgoals (e.g., ``identify relevant quantities,'' ``formulate equations,'' ``solve subexpressions''), which guide the model through a structured solution path. 

\begin{figure*}
    \centering
    \includegraphics[width=\linewidth]{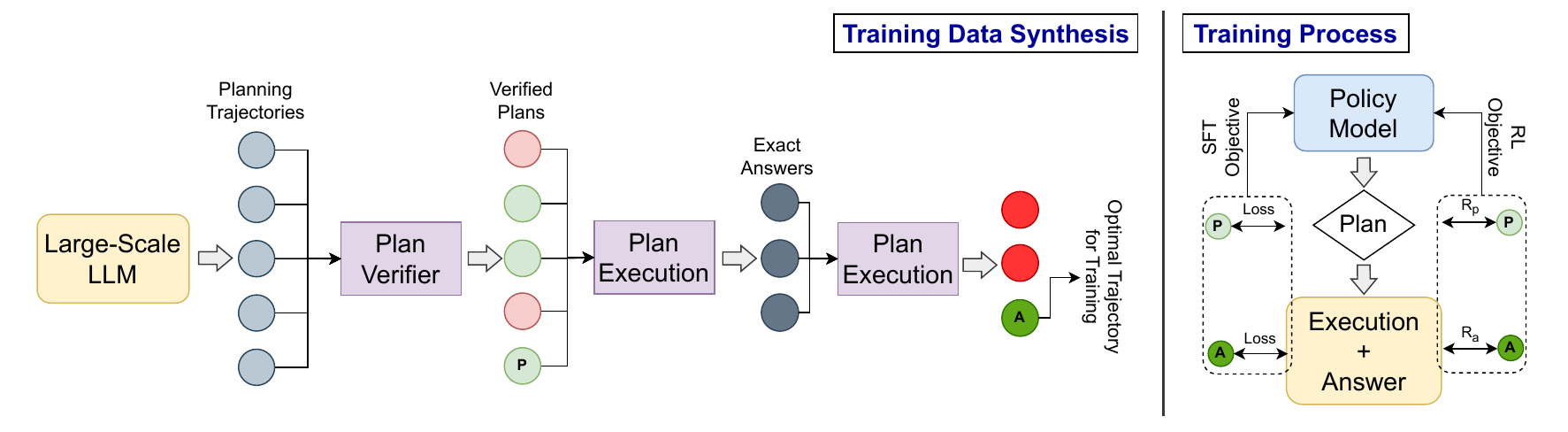}
    \caption{\textbf{Left:} Large-scale LLM generates multiple planning candidates (green = high-quality, red = low-quality). A \textit{Plan Verifier} scores plans, and an \textit{Answer Verifier} confirms the final answer; only trajectories passing both become the training corpus. \textbf{Right:} These trajectories train the policy model via SFT and RL objectives. P and A denote gold synthetic planning and answer trajectories, while $R_p$ and $R_a$ are their respective rewards.}
    \label{fig:method_1}
\end{figure*}

However, these capabilities have largely been explored at inference time and on large-scale proprietary models \cite{wang2planning, parmar2025plangen}; smaller open-source LLMs still struggle to leverage the decomposition step from natural planning effectively, limiting their performance on complex tasks. Thus, we address the research question: ``\textbf{Can we improve the performance of smaller LLMs on complex tasks by incorporating such capabilities through post-training, rather than relying solely on inference-time prompts?}'' To this end, we propose \textsc{Plan-Tuning}, a post-training method that uses synthetic planning trajectories—sequences of natural decomposition steps—to teach a model how to plan as part of its parameterized knowledge. \textsc{Plan-Tuning} incorporates planning trajectories in supervised and Reinforcement Learning (RL) settings, with customized objectives (loss) and reward functions to improve planning capabilities. Our plan-tuned smaller LLMs show improved reasoning skills, focused on mathematical reasoning.

% \begin{figure}
%     \centering
%     \includegraphics[width=0.82\linewidth]{figures/method_1.pdf}
%     \caption{\textbf{Top (Data Synthesis):} Large-scale LLM generates multiple planning candidates (green = high-quality, red = low-quality). A \textit{Plan Verifier} score plans, and an \textit{Answer Verifier} confirms final answer; only trajectories passing both become the training corpus. \textbf{Bottom (Training):} These curated trajectories train the policy model via SFT and GRPO, incorporating planning capabilities in smaller LLMs.}
%     \label{fig:method_1}
% \end{figure}

For distilling high-quality planning trajectories from large-scale LLM, we leverage the Best-of-$\mathcal{N}$ approach from the recent PlanGEN framework proposed by \citet{parmar2025plangen}. For the MATH and GSM8k \cite{cobbe2021training} training sets, we generate five candidate plans per problem. Each plan is then evaluated by the verification agent from PlanGEN, and only those exceeding a predefined quality threshold are retained. Because gold final answers are available for training data, we next execute every retained plan and verify that it produces the correct final answer. At last, we include only those trajectories in \textsc{Plan-Tuning} that both pass the agent-based scoring threshold and yield the correct solution; all others are discarded. The whole process is illustrated in Figure \ref{fig:method_1}.

In \textsc{Plan-Tuning}, we explore two post-training paradigms, supervised fine-tuning (SFT) and RL to incorporate planning in smaller LLMs. In SFT, we examine (1) an end-to-end setting where the model learns to map problem statements directly to a step-by-step plan and final answer, and (2) a two-stage pipeline that first generates only the plan and then executes it to derive the solution. In RL, we introduce Group Relative Policy Optimization (GRPO) \cite{shao2024deepseekmath}, in which we augment the RL objective with planning-specific rewards to incorporate high-quality plan generation. Both Proximal Policy Optimization (PPO) \cite{ouyang2022training} and Direct Preference Optimization (DPO) \cite{rafailov2023direct} depend on human‐annotated or model‐annotated preference data to learn a reward signal, which requires a separate data‐collection or synthesis pipeline, but our work do not focus on synthesizing data for preference optimization. By contrast, GRPO leverages our existing LLM‐based plan verifier to generate rewards without extra annotation effort. Furthermore, recent studies \cite{shao2024deepseekmath} have demonstrated that GRPO outperforms PPO for mathematical reasoning tasks. For these reasons, we selected the SOTA method, GRPO, in our preliminary study. We evaluate \textsc{plan-tuning} against two strong baselines: an SFT model trained using reasoning chains and answers, and a vanilla GRPO model that optimizes preferences without any planning-based reward.  

We evaluate four mathematical reasoning benchmarks—two in-domain (GSM8k and MATH) for both training and evaluation, and two out-of-domain (OlympiadBench \cite{he2024olympiadbench} and AIME \cite{sun2025challenging}) to assess generalization. For SFT, we use Gemma-3-12B-it \cite{team2025gemma} and Qwen3-8B \cite{qwen2.5}, and for RL, we use Gemma-3-1B-it, and Qwen3-4B. Across both in-domain tasks, plan-tuned models consistently outperform these baselines, yielding $\sim7\%\uparrow$ on GSM8k and $\sim20\%\uparrow$ on MATH. They also demonstrate stronger reasoning-based generalization, achieving $\sim10\%\uparrow$ and $\sim12\%\uparrow$ on OlympiadBench and AIME, respectively. Moreover, our detailed analysis results in several interesting findings, such as mixing GSM8k and MATH during plan-tuning actually degrades performance, and plan-tuned models substantially improve performance on longer reasoning chains. In summary, our work proposes \textsc{Plan-Tuning}, a method to distill planning trajectories from large-scale LLMs and fine-tune smaller LLMs to incorporate this ability to solve complex problems. We believe that \textsc{Plan-Tuning} can be an effective method for improving complex reasoning in smaller LLMs.

\section{Related Works}
\label{sec:related_work}

Decomposing complex problems into easy and smaller sub-tasks has been proven effective to improve performance of LLMs. Many prompt-level methods \cite{liu2023pre, patel2022question, kuznia2022less} has been proposed to this end. For instance, Least-to-Most \cite{zhouleast} prompting iteratively breaks down a complex problem into a series of simpler subproblems and then solve them in sequence. Similarly, Plan-and-Solve \cite{wang2023plan} prompting consists of two components: first, devising a plan to divide the entire task into smaller subtasks, and then carrying out the subtasks according to the plan. In contrast to these prompt-level approaches, we aim to achieve this decomposition ability via post-training of small-scale LLMs at inference-time. Thus, we present related literature in these two areas: (1) Inference-time scaling methods for LLMs, and (2) Post-training techniques for task decomposition.

\paragraph{Inference-time Scaling in LLMs}
Inference-time algorithms have recently shown a powerful way to optimize LLM output during inference, providing significant improvements in accuracy without scaling the model. Chain-of-thought (CoT) prompting and its variants \citep{wei2022chain, kojima2022large} showed that adding intermediate reasoning steps during inference-time greatly boosts the performance of LLMs. New methods have been proposed, such as self-consistency \citep{wang2022self}, which generates multiple reasoning chains from LLM and then selects the final answer based on majority voting. One very popular approach is the use of Monte Carlo Tree Search (MCTS) \cite{zhang2024accessing}, which iteratively explores multiple solution paths during inference. This technique has been successfully applied to models like LLaMa-3-8B, which integrates a self-refinement mechanism that allows the model to revisit and improve its initial solutions over time. Another method, test-time optimization \cite{snell2024scaling}, focuses on dynamically adjusting computational resources during inference. By optimizing compute resources based on the complexity of a task, this approach strikes a balance between efficiency and accuracy, ensuring that difficult tasks receive more attention while simpler tasks are processed with fewer resources. Additionally, compute-optimal inference \cite{wu2024empirical} highlights the importance of effectively distributing computational power during problem-solving tasks. Finally, repeated sampling \cite{brown2024large} is a technique that uses multiple inference attempts to improve solution quality. \citet{wang2planning} uses the inference time algorithms to improve LLMs planning capabilities to solve code synthesis problems. Recent works such as \citet{parmar2025plangen} show that better natural language planning improves downstream reasoning capabilities of underlying LLMs. In contrast to all these approaches focused on increasing compute at inference-time, our work focuses on inference-aware post-training with planning trajectories.  

\paragraph{Post-training Methods for LLMs}
Past attempts have been made to improve the performance of smaller LMs using various types of training trajectories \cite{ouyang2022training, rafailov2023direct, saeidi2024triple}. Our work is closer to \citet{jiao-etal-2024-learning} where authors propose a two-stage approach: first, creating human-like, step-by-step planning trajectories, then automatically synthesizing detailed ``process rewards'' that score how accurately each trajectory follows its plan. In a similar direction, \citet{song-etal-2024-trial} pioneers an exploration-driven paradigm, generating multiple candidate trajectories and iteratively refining them via performance feedback to improve long-horizon reasoning. \citet{zhang2024agent} proposes a pipeline that unifies demonstration data, self-play rollouts, and human preferences, demonstrating that diverse training signals yield more robust agent policies. \citet{wang2024learning} shows that incorporating explicitly negative trajectories during fine-tuning helps the model learn to avoid erroneous or unsafe actions. Other work investigates curriculum design and data curation strategies, revealing that carefully scheduled exposure to increasingly complex tasks significantly boosts final policy quality \cite{chen-etal-2024-agent}. Meanwhile, \citet{song-etal-2024-agentbank} demonstrates that scaling up trajectory volume and diversity is key to robust generalization. Unlike prior works that apply planning only at inference or via large-scale behavior cloning, \textsc{Plan-Tuning} incorporates synthetic planning trajectories into small LLMs’ parameters through supervised and RL post-training objectives.

\section{Proposed Method}
\label{sec:methods}

\subsection{Task Formulation}

\paragraph{Defining Task Decomposition as Part of Natural Planning} 

%Here, we are defining formally how task decomposition happens as part of natural planning for the scope of this work, and we are not focusing on traditional AI planning from the literature. In \textsc{Plan-Tuning}, we define planning as the process of decomposing a problem $x$ into a sequence of intermediate subgoals that systematically guide the solution. 

In this work, we formalize natural planning—distinct from classical AI planning (which is not the focus of this work)—as the process of decomposing an input problem $x$ into an ordered sequence of intermediate subgoals that guide its solution. Formally, we introduce a latent state space $\mathcal{S}$ of partial reasoning states, with the initial state $s_0 = x$. A planning trajectory $\tau = (\tau_1,\dots,\tau_K)$ is then defined as a sequence of operators $\tau_k: \mathcal{S}\to\mathcal{S}$ such that

\begin{equation}
s_k \;=\; \tau_k(s_{k-1})\quad\text{for }k=1,\dots,K,
\end{equation}

The final state $s_K$ encodes sufficient information to extract the correct answer $y$. We model the planner as a conditional distribution as below:

\begin{equation}
\pi_\theta(\tau\mid x)\;=\;\prod_{k=1}^K \pi_\theta\bigl(\tau_k \mid s_{k-1}\bigr),
\end{equation}

This equation indicates that the policy model $\pi_\theta$ assigns high probability to trajectories that (i) follow high-quality, human-like decomposition patterns and (ii) lead to a correct final solution. For the scope of this work, we use smaller LLMs as policy models for generating planning trajectories.

\begin{figure*}
    \centering
    \includegraphics[width=\textwidth]{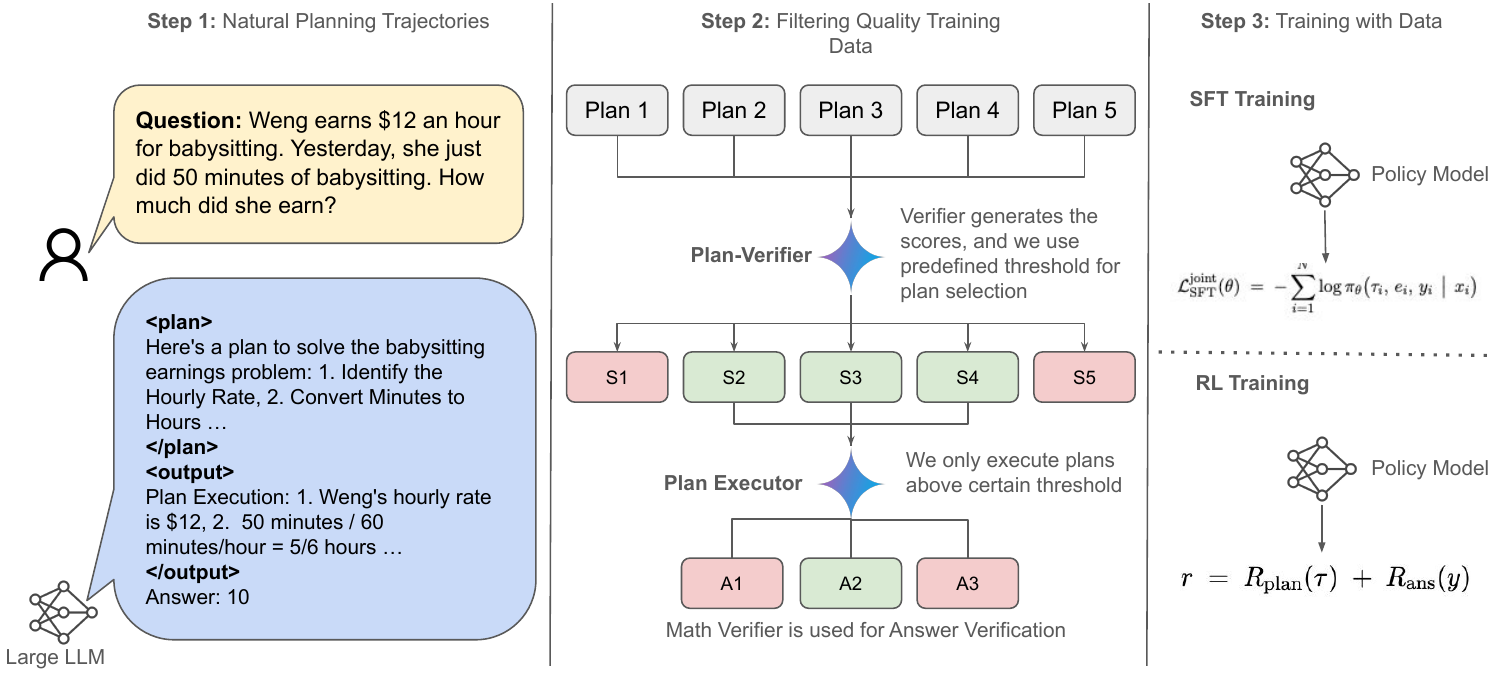}
    \caption{Overview of the \textsc{Plan-Tuning} pipeline. First, a large LLM generates multiple candidate natural-language planning trajectories for each problem. Next, a Plan Verifier scores and filters these trajectories, and a Math Verifier executes and validates only those above a quality threshold. Finally, the curated plan–answer pairs are used to train the target model via both SFT and RL (GRPO) objectives.}
    \label{fig:method_2}
\end{figure*}

\paragraph{Problem Statement} We frame mathematical reasoning with planning as learning a conditional model that, given a problem statement $x\in\mathcal{X}$, jointly generates a planning trajectory $\tau=(\tau_1,\dots,\tau_K)\in\mathcal{T}$ and a final answer $y\in\mathcal{Y}$, as shown in the below equation.

\begin{equation}
\pi_\theta(\tau,y\mid x) \;=\;\pi_\theta(y\mid \tau,x)\,\pi_\theta(\tau\mid x)
\end{equation}

The above equation is the policy of our plan-tuned LLM, parameterized by $\theta$. We collect a training corpus $\mathcal{D}=\{(x_i,\tau_i,y_i)\}_{i=1}^N$ of synthetic, high-quality trajectories paired with gold answers. For this purpose, we use large-scale LLMs such as Gemini-2.0-Flash along with a filtering module to synthesize high-quality data. Our learning process thus train model to first decompose problem into solvable subproblems through natural planning, and leverage two different objectives: supervised fine-tuning, which aligns $\pi_\theta(\tau,y\mid x)$ with the observed $(\tau_i,y_i)$ pairs, and reinforcement learning, which maximizes expected rewards to encourage better generation, execution and correct answers.

\subsection{Data Generation}

Figure \ref{fig:example} and Figure \ref{fig:method_2} show illustrative examples of planning trajectories and their execution to get the final answer for the given problem\footnote{More examples are provided at \url{https://github.com/Mihir3009/plan-tuning}}. Here, we provide a detailed discussion about the distillation of these high-quality planning trajectories using Gemini-2.0-Flash. For this purpose, we utilize the training sets of GSM-8k and MATH.

\paragraph{Data Synthesis} 
Let us denote the training set as $\{x_i\}_{i=1}^N$. Motivated by \citet{parmar2025plangen}, for each problem $x_i$, we employ a method similar to the PlanGEN (Best of $\mathcal{N}$) (with $\mathcal{N}=5$ and temperature to 0.7 for the underlying LLM $\mathcal{M}$ for diversity) to synthesize five distinct natural-language planning trajectories:

\begin{equation}
\{\tau_i^{(1)}, \ldots, \tau_i^{(5)}\}\;\sim\;\mathcal{M}(\,\cdot\mid x_i).
\end{equation}

Each trajectory $\tau_i^{(n)}$ is passed through constraint-based verification agent---adapted from \citet{parmar2025plangen}---to compute a plan quality score.

\begin{equation}
s_i^{(n)}=R_{\mathrm{ver}}(\tau_i^{(n)}),
\end{equation}

This score assesses coherence, logical soundness, and alignment with human-like decomposition patterns. Simultaneously, we execute each $\tau_i^{(n)}$ by feeding it into our execution module (another underlying LLM) to obtain an answer $y_i^{(n)}$. This yields a candidate set as below, capturing both plan quality and solution correctness.

\[
\bigl\{\,x_i,\;\tau_i^{(n)},\;y_i^{(n)},\;s_i^{(n)}\bigr\}_{
\substack{i=1,\dots,N \\ n=1,\dots,5}
},
\]

The whole process of doing this data synthesis is presented in Figure \ref{fig:method_1}, and a detailed example is presented in Figure \ref{fig:method_2}. Also, prompts used for this data synthesis method are provided in App. \ref{app:data_prompt}.
 
\paragraph{Training Data Quality}
To ensure high-quality supervision, we apply a two-stage filtering process. First, we retain only trajectories whose verification score exceeds a threshold $\alpha=80$:

\begin{equation}
s_i^{(n)}\;\ge\;80,
\end{equation}

This threshold we decided based on manual analysis presented in \citet{parmar2025plangen} that plans above this score have a high likelihood of yielding correct solutions. Now, for the training set, we have the gold final answer available. Hence, we validate each selected trajectory by checking execution correctness, i.e., $y_i^{(n)} = y_i^*$, where $y_i^*$ is the gold answer. Only those $(x_i,\tau_i^{(n)},y_i^{(n)})$ triples satisfying both criteria are included in the final training corpus; all remaining candidates are discarded. This selection yields a dataset of \texttt{<problem, plan, plan execution, final answer>} that balances plan quality with solution accuracy, creating high-quality training data.

\subsection{\textsc{plan-tuning}}

We utilize two post-training methods: (i) supervised fine-tuning (SFT), and (ii) reinforcement learning via Gradient-based Reward Policy Optimization (GRPO)—each aimed to incorporate planning abilities in underlying LLMs.

\subsubsection{\textsc{plan-tuning}: SFT Training}

In SFT, we compare two approaches: (1) joint plan and answer Generation teaches models to produce complete solutions (plans, step-by-step execution, and final answers) from problems, while (2) plan-only generation focuses exclusively on creating high-quality plans. These methods use different loss functions to optimize model parameters, with the joint approach minimizing negative log-likelihood across all solution components and the plan-only approach focusing only on plan quality.

\paragraph{Method 1} 
The model learns to map each problem $x_i$ to the concatenated sequence $(\tau_i, e_i, y_i)$, where $\tau_i$ is the plan, $e_i$ is the step-by-step execution of that plan, and $y_i$ is the final answer. We minimize the negative log-likelihood using below:

\begin{equation}
\mathcal{L}_{\mathrm{SFT}}^{\mathrm{joint}}(\theta)
  \;=\;
  -\sum_{i=1}^{N}
    \log \pi_{\theta}\bigl(\tau_i,\,e_i,\,y_i \,\bigm|\, x_i\bigr)\,
\end{equation}

\paragraph{Method 2} 
The model focuses exclusively on generating high-quality plans $\tau_i$. The objective we use to optimize is given below:

\begin{equation}
\mathcal{L}_{\mathrm{SFT}}^{\mathrm{plan}}(\theta)
  \;=\;
  -\sum_{i=1}^{N}
    \log \pi_{\theta}\bigl(\tau_i \,\bigm|\, x_i\bigr)\,
\end{equation}

Once we have generated a high-quality plan $\tau_i$, we use the same off-the-shelf base LLM to execute that plan and produce the final answer. No additional training is performed on this execution model. In the rest of the paper, we refer to method 1 as $\mathcal{M}_1$, and method 2 as $\mathcal{M}_2$.

\subsubsection{\textsc{plan-tuning}: GRPO Training}

In this approach, we apply GRPO, a policy-gradient algorithm that directly maximizes sequence-level returns. Let $q$ denote a problem statement and $o=(o_1,\dots,o_{|o|})$ the model's generated output/rollouts (the plan, its execution, and the final answer). We define a combined reward for each sampled trajectory as below:

\begin{equation}
r(.) \;=\; R_{\mathrm{plan}}(\tau)\;+\;R_{\mathrm{ans}}(y),
\label{eq:reward}
\end{equation}
where $R_{\mathrm{plan}}(\tau)$ is the plan-quality score produced by our similarity function (discussed later in the section), and $R_{\mathrm{ans}}(y)$ is a binary correctness indicator ($2$ if $y$ matches the gold answer, $0$ otherwise).

\paragraph{Background} GRPO, a PPO \cite{schulman2017proximal} variant, estimates the advantage by aggregating reward scores of a group of \(n\) sampled responses to a given query $q$. Formally, let $\pi_{\theta}$ and $\pi_{\theta_{old}}$ be the current and old policy models respectively. Let $q$ and $o_{i}$ be the query and $i^{th}$ response sampled from the dataset and the old policy, respectively. Let $r(.)$ be the reward function, which measures the correctness of a given response. Then, the GRPO objective is defined as follows: \\

\resizebox{0.9\linewidth}{!}{$%
\begin{aligned}
\mathcal{J}_{\rm GRPO}(\theta)&= \mathbb{E}\Bigl[\,
q\sim \mathcal{D},\;\{o_i\}_{i=1}^{n}\sim \pi_{\theta_{\rm old}}(O\mid q)
\Bigr] \\[-0.4ex]
&\quad\Biggl\{
\frac{1}{n}\sum_{i=1}^n \frac{1}{|o_i|}
\sum_{t=1}^{|o_i|}
\min\Bigl[
\tfrac{\pi_\theta(o_{i,t}\mid q,o_{i,<t})}
{\pi_{\theta_{\rm old}}(o_{i,t}\mid q,o_{i,<t})}\,\hat A_{i,t},\\
&\operatorname{clip}\bigl(\tfrac{\pi_\theta(o_{i,t}\mid q,o_{i,<t})}
{\pi_{\theta_{\rm old}}(o_{i,t}\mid q,o_{i,<t})},
1-\epsilon,\,1+\epsilon\bigr)\,\hat A_{i,t}
\Bigr] \\
&- \beta\,D_{KL}\bigl[\pi_\theta\bigm\|\pi_{\rm ref}\bigr]
\Biggr\}\
\end{aligned}
$}

Here, the advantage is calculated as the normalized reward, i.e., $\hat A_{i,t} = \tilde r(o_i) =\frac{r(o_i) - \operatorname{mean}(r)}{\operatorname{std}(r)}$. This $\hat A_{i,t}$ centers and scales each trajectory's return $r$ across the sampled batch. By weighting each token's log-probability gradient by $\hat A$, GRPO amplifies updates for outputs yielding above-average combined reward and suppresses those with below-average reward.

\paragraph{Our Modification (Plan GRPO)} In the above formulation, our focus is to modify  $r(.)$ function, and incorporate planning-based reward. So, the final reward function consists: (1) plan-quality reward $R_{\mathrm{plan}}(\tau)$ ensures that generated trajectories closely match high-quality synthetic plans distilled from large-scale LLMs, and (2) answer correctness reward $R_{\mathrm{ans}}(y)$ guarantees that following these plans leads to right solutions. Thus, the final reward is presented in Equation \ref{eq:reward}. By normalizing and combining these signals, GRPO fine-tunes the policy model $\pi_\theta$ to generate both good plans, accurate execution, and a correct final answer.

% \paragraph{Background} GRPO estimates the gradient of the objective $\mathcal J(\theta)$ with the classic PPO update, normalized by sequence length and reward variance:

% \begin{equation}
% \begin{aligned}
% \nabla_\theta \mathcal{J}(\theta) 
% = \mathbb{E}_{(q,o)\sim\mathcal D}\!
% \Biggl[\, \frac{1}{|o|}\sum_{t=1}^{|o|} 
% \underbrace{\hat{A}}_{\displaystyle\frac{r - \overline{r}}{\mathrm{std}(r)}} \\
% \;\times\; \nabla_\theta \log \pi_\theta(o_t \mid q, o_{<t})
% \Biggr].
% \end{aligned}
% \end{equation}
% \begin{equation}
% \begin{aligned}
% \nabla_\theta \mathcal{J}(\theta) 
% = \mathbb{E}_{(q,o)\sim\mathcal D}\!
% \Biggl[\, \frac{1}{|o|}\sum_{t=1}^{|o|} 
% \hat{A} \\
% \;\times\; \nabla_\theta \log \pi_\theta(o_t \mid q, o_{<t})
% \Biggr].
% \end{aligned}
% \end{equation}
% \text{Here,}
% \begin{equation*}
% \hat{A} = \frac{r - \mathrm{mean}(r)}{\mathrm{std}(r)}
% \end{equation*}

% $\hat A$ is the normalized advantage, which centers and scales each trajectory's return $r$ across the sampled batch. By weighting each token's log-probability gradient by $\hat A$, GRPO amplifies updates for outputs yielding above-average combined reward and suppresses those with below-average reward.

% Equivalently, we can express the GRPO training objective as minimizing the loss as below:

% \begin{equation}
% \begin{aligned}
% \mathcal{L}_{\mathrm{GRPO}}(\theta) = -\mathbb{E}_{(q,o)\sim\mathcal D} \Bigl[ 
% \hat{A} \times \frac{1}{|o|} \\
% \times \sum_{t=1}^{|o|} \log \pi_\theta(o_t|q, o_{<t})
% \Bigr]
% \end{aligned}
% \label{eq:grpo}
% \end{equation}

\paragraph{Details on Planning Reward $R_{\mathrm{plan}}(\tau)$} is computed by measuring how closely a model-generated plan $\tau$ matches its reference $\tau^*$ using our Gemini-based similarity scorer. Here, reference $\tau^*$ is the plan synthesized from above section. It is distilled from large-scale LLMs (Gemini-2.0 in our case). We first extract the plan segments. We parse each model completion to pull out only the \texttt{<plan>} portion.
Then, we prompt Gemini for similarity. We construct a natural-language prompt comparing the generated plan to the gold plan and send it to Gemini-2.0-Flash, asking it to rate plan similarity on a 0--1 scale. At last, we parse the numeric score. We apply a flexible regular expression to the returned text to extract the score.

\begin{equation*}
R_{\mathrm{plan}}(\tau_i)\;=\;\text{Score}_{\mathrm{Gemini}}\bigl(\tau_i,\tau_i^*\bigr)\;\in\;[0,1].
\end{equation*}

These per-example rewards are then fed into our GRPO objective $\mathcal{J}_{\rm GRPO}$, so that higher-quality plans receive proportionally larger policy-gradient updates. The prompt for calculating $R_{\mathrm{plan}}(\tau)$ is presented in App. \ref{app:planning_reward}.

\section{Results and Analysis}
\label{sec:results}

\subsection{Experimental Setup}

\begin{table}
\centering
\footnotesize
\resizebox{0.75\linewidth}{!}{
\begin{tabular}{c|c|c}
\toprule
\textbf{Dataset}      & \textbf{Train} & \textbf{Eval} \\
\midrule
GSM-8k                & 6,586          & 1,319         \\
MATH                  & 10,000         & 500           \\
OlympiadBench         & --             & 674           \\
AIME                  & --             & 933           \\
\bottomrule
\end{tabular}}
\caption{Statistics of the datasets used in our experiments. Training set sizes are shown for in-domain benchmarks, and evaluation set sizes for both in-domain and out-of-domain benchmarks.}
\label{tab:dataset_stats}
\end{table}

% \usepackage{multirow}
\begin{table*}
\centering
\resizebox{\textwidth}{!}{
\begin{tabular}{c|c|ccc|ccc}
\toprule
\multirow{2}{*}{\textbf{Models}}  & \multirow{2}{*}{\textbf{Methods}} & \multicolumn{3}{c|}{\textbf{Datasets Evaluated using $\mathcal{M}$(GSM8k)}}                                                                                         & \multicolumn{3}{c}{\textbf{Datasets evaluated using $\mathcal{M}$(MATH)}}                                                                                         \\ \cmidrule{3-8} 
                                  &                                   & \multicolumn{1}{c|}{\textbf{GSM8k}} & \multicolumn{1}{c|}{\textbf{\begin{tabular}[c]{@{}c@{}}OlympiadBench\\ (MATH)\end{tabular}}} & \textbf{AIME 2024} & \multicolumn{1}{c|}{\textbf{MATH}} & \multicolumn{1}{c|}{\textbf{\begin{tabular}[c]{@{}c@{}}OlympiadBench\\ (MATH)\end{tabular}}} & \textbf{AIME 2024} \\ \midrule
\multirow{5}{*}{\textbf{Gemma-3}} & Baseline SFT                      & \multicolumn{1}{c|}{81.43}          & \multicolumn{1}{c|}{31.45}                                                                   & 28.62              & \multicolumn{1}{c|}{65.40}          & \multicolumn{1}{c|}{22.11}                                                                   & 14.04              \\ 
                                  & Vanilla GRPO                      & \multicolumn{1}{c|}{19.94}              & \multicolumn{1}{c|}{02.52}                                                                       & 00.00                  & \multicolumn{1}{c|}{\cellcolor{gray!15}}             & \multicolumn{1}{c|}{\cellcolor{gray!15}}                                                                       & \cellcolor{gray!15}                  \\ \cmidrule{2-8} \cmidrule{2-8} 
                                  & $\mathcal{M}_1$                          & \multicolumn{1}{c|}{86.20}           & \multicolumn{1}{c|}{32.20}                                                                    & 32.27              & \multicolumn{1}{c|}{74.20}          & \multicolumn{1}{c|}{27.00}                                                                      & 20.04              \\  
                                  & $\mathcal{M}_2$                          & \multicolumn{1}{c|}{\textbf{86.35}}          & \multicolumn{1}{c|}{\textbf{34.27}}                                                                   & \textbf{33.98}              & \multicolumn{1}{c|}{\textbf{83.80}}          & \multicolumn{1}{c|}{\textbf{31.75}}                                                                   & \textbf{29.37}              \\ 
                                  & Plan GRPO                         & \multicolumn{1}{c|}{\textbf{28.35}}              & \multicolumn{1}{c|}{\textbf{04.30}}                                                                       & \textbf{03.00}                  & \multicolumn{1}{c|}{\cellcolor{gray!15}}             & \multicolumn{1}{c|}{\cellcolor{gray!15}}                                                                       & \cellcolor{gray!15}                  \\ \midrule \midrule
\multirow{5}{*}{\textbf{Qwen-3}}  & Baseline SFT                      & \multicolumn{1}{c|}{80.74}          & \multicolumn{1}{c|}{28.78}                                                                   & 29.05              & \multicolumn{1}{c|}{53.20}          & \multicolumn{1}{c|}{12.02}                                                                   & 05.57               \\ 
                                  & Vanilla GRPO                      & \multicolumn{1}{c|}{84.27}          & \multicolumn{1}{c|}{24.63}                                                                       & \textbf{17.68}                  & \multicolumn{1}{c|}{\cellcolor{gray!15}}             & \multicolumn{1}{c|}{\cellcolor{gray!15}}                                                                       & \cellcolor{gray!15}                  \\ \cmidrule{2-8} \cmidrule{2-8}
                                  &  $\mathcal{M}_1$                       & \multicolumn{1}{c|}{\textbf{87.11}}          & \multicolumn{1}{c|}{\textbf{30.27}}                                                                   & \textbf{32.05}              & \multicolumn{1}{c|}{73.80}          & \multicolumn{1}{c|}{23.44}                                                                   & 19.08              \\ 
                                  & $\mathcal{M}_2$                          & \multicolumn{1}{c|}{85.67}          & \multicolumn{1}{c|}{30.12}                                                                   & 31.94              & \multicolumn{1}{c|}{\textbf{79.40}}          & \multicolumn{1}{c|}{\textbf{31.01}}                                                                   & \textbf{31.51}              \\  
                                  & Plan GRPO                         & \multicolumn{1}{c|}{\textbf{86.57}}          & \multicolumn{1}{c|}{\textbf{25.07}}                                                                       & 15.22                  & \multicolumn{1}{c|}{\cellcolor{gray!15}}             & \multicolumn{1}{c|}{\cellcolor{gray!15}}                                                                       & \cellcolor{gray!15}                  \\ \bottomrule
\end{tabular}
}
\caption{This table reports the accuracy (\%) of the base LLM and its two plan-tuned variants (custom SFT and GRPO) on four mathematical reasoning benchmarks. Columns 1–2 show in-domain performance on GSM8K and MATH. Columns 3–4 present out-of-domain generalization on OlympiadBench and AIME 2024. These results demonstrate that leveraging synthetic planning trajectories via both SFT and RL objectives improves reasoning accuracy in smaller LLMs. $\mathcal{M}$ (GSM8k): Model trained using GSM8k dataset, $\mathcal{M}$ (MATH): Similar for MATH.}
\label{tab:main_results}
\end{table*}

\paragraph{Datasets}
We evaluate our \textsc{Plan-Tuning} on four mathematical reasoning benchmarks. As shown in Table \ref{tab:dataset_stats}, GSM-8k \cite{cobbe2021training} and MATH \cite{hendrycks2021measuring} provide in-domain datasets for both training and evaluation. After filtering out lower-quality examples, the GSM-8k training set contains 6,586 problems (from an original 7,500), with 1,319 held out for evaluation. The MATH training set comprises 10,000 problems (from 12,000), with 500 held-out for evaluation. To evaluate out-of-domain generalization, we use the text-only version of OlympiadBench (MATH) (674 problems) \cite{he2024olympiadbench} and AIME (933 problems), for only evaluation purposes.

\paragraph{Models}
For data synthesis, we use \textbf{Gemini-2.0-Flash} (Checkpoint: April 2025). We fine-tune four pretrained HuggingFace checkpoints: \textbf{Gemma-3-12B-It} and \textbf{Qwen3-8B} for \textsc{plan-tuning}: SFT; and \textbf{Gemma-3-1B-It} and \textbf{Qwen3-4B} for \textsc{plan-tuning}: GRPO.

\paragraph{Baselines}
Our two baselines for these training paradigms: (i) an SFT model that learns to output conventional chain-of-thought reasoning and final answers, and (ii) a ``vanilla'' GRPO model optimized only on answer correctness $R_{\mathrm{ans}}(y)$ without any planning-specific rewards $R_{\mathrm{plan}}(\tau)$.

\paragraph{Proposed Experiments}
For \textsc{plan-tuning}: SFT, we experiment with both methods: (i) $\mathcal{M}_1$, a joint plan-and-answer formulation, where the model maps each input $x$ to the tuple $\langle\text{Plan},\;\text{Execution},\;\text{Answer}\rangle$, and (ii) $\mathcal{M}_2$, a plan-only variant in which the model simply generates $\langle\text{Plan}\rangle$. Both SFT variants use a batch size of 8, an adaptive learning rate of $5\times10^{-6}$, a single training epoch, and a cosine learning-rate scheduler. Second, we apply GRPO Training: a vanilla GRPO baseline that optimizes only the answer correctness reward $R_{\mathrm{ans}}(y)$, and a planning-specific GRPO that additionally incorporates the Gemini-based planning reward $R_{\mathrm{plan}}(\tau)$ into the objective $R_{\mathrm{plan}} + R_{\mathrm{ans}}$. For GRPO, we use a batch size of 32, the same learning rate and scheduler as SFT, one epoch, 4 rollouts per policy update, and a KL-coefficient of $0.04$. Due to resource and time constraints, our GRPO experiments are limited to the GSM8k dataset.  

\paragraph{Metrics}
We report dataset-specific accuracy on each benchmark to assess in-domain performance and out-of-domain generalization. In particular, we use micro-average accuracy for OlympiadBench similar to \citet{he2024olympiadbench}, and Exact Match (EM) for all other datasets. 

\subsection{Main Results}

% In this section, we describe the results from Table \ref{tab:main_results} in detail.

\paragraph{Baseline Performance on In-Domain Tasks}
From Table \ref{tab:main_results}, the off-the-shelf SFT model trained on GSM8K achieves 81.43\% accuracy when evaluated on GSM8K and 65.4\% on the MATH benchmark for the Gemma-3. In comparison, the Qwen-3 SFT baseline shows 80.74\% on GSM8K and only 53.2\% on MATH. This is a $\sim15\%$ drop between the two in-domain datasets, highlighting that, without explicit natural planning, smaller LLMs struggle to generalize from more complex and constrained math word problems where the diverse, multi-step reasoning is required.

% Comparing with SFT baselines, $\mathcal{M}_1$ and $\mathcal{M}_2$ of \textsc{plan-tuning} consistently outperform by a large margin ($\sim7\%\uparrow$ GSM8k, and $\sim23\%\uparrow$ MATH). This large $\uparrow$ in MATH highlights that, without explicit natural planning, smaller LLMs struggle to generalize from more complex and constrained math word problems where the diverse, multi-step reasoning is required.

\paragraph{Supervised Fine-Tuning Variants}
Introducing supervised planning trajectories yields consistent gains across both model families. For Gemma-3, the joint plan-and-answer SFT ($\mathcal{M}_1$) improves GSM8K accuracy to 86.2\% ($+4.8\%$) and MATH to 74.2\% ($+8.8\%$), while the plan-only SFT ($\mathcal{M}_2$) further boosts these to 86.35\% on GSM8K and an 83.8\% on MATH. Qwen-3 exhibits similar trends: joint SFT improves up to 87.11\% on GSM8K ($+6.4\%$) and 73.8\% on MATH ($+20.6\%$), whereas plan-only fine-tuning yields 85.67\% and 79.4\%, respectively. In particular, large gains on MATH suggest that guiding the model to focus purely on plan generation better incorporates the structured decomposition strategies needed for complex, multi-step reasoning tasks.

\paragraph{Out-of-Domain Generalization}
When evaluated on Olympiad-level math benchmarks, the importance of \textsc{plan-tuning} is even more prominent. Gemma-3 baseline SFT achieves only 31.45\% on OlympiadBench and 14.04\% on AIME. Joint SFT (Method 1) improves it to 32.2\% and 20.04\%; plan-only SFT (Method 2) improves them further to 34.27\% and 29.37\%. Qwen-3 follows the same trend: baseline SFT is 28.78\%/5.57\% (OlympiadBench/AIME), joint SFT 30.27\%/19.08\%, and plan-only 30.12\%/31.51\%. These gains, often improving twice or more AIME performance, demonstrate that high-quality planning exemplars for training are especially critical for tackling novel, complex Olympiad-level math problems. 

\paragraph{Improvements with Plan-GRPO}
From Table \ref{tab:main_results}, we present results for vanilla-GRPO and Plan-GRPO where the model is trained using the GSM8k dataset. From the results, we can observe that, for Gemma-3-1B-It, plan-GRPO improves performance on GSM8k to 28.35\% compared to vanilla-GRPO (19.94\%). A similar trend can be observed for the Qwen3-4B model in terms of GSM8k. Now, on out-of-domain datasets, for Gemma-3, plan-tuned models are achieving better generalization. However, for Qwen-3 models, we are seeing a performance drop for AIME. Training with GRPO is highly dependent on reward functions. Also, the lower performance on OlympiadBench and AIME is subject to the smaller sizes of LLMs, 1B and 4B, used for GRPO training.

\paragraph{Lower performance of GRPO is attributed to model-size.} Our GRPO experiments were conducted on significantly smaller models (Gemma-3-1B-It and Qwen3-4B) due to resource constraints, whereas SFT experiments used larger models (Gemma-3-12B-It and Qwen3-8B). This size gap likely contributed to the performance drop, especially given that RL-based training is more sensitive to model capacity for stable policy optimization. To assess the impact of model size, we conducted an additional case study during the limited rebuttal period, where we trained Gemma-3-1B-It using SFT on GSM8K. The resulting accuracy is $5.16\%$, which is even lower than the results on vanilla-GRPO ($19.94\%$). This supports that model size is a major factor in the performance gap.

\begin{figure}
    \centering
    \includegraphics[width=\linewidth]{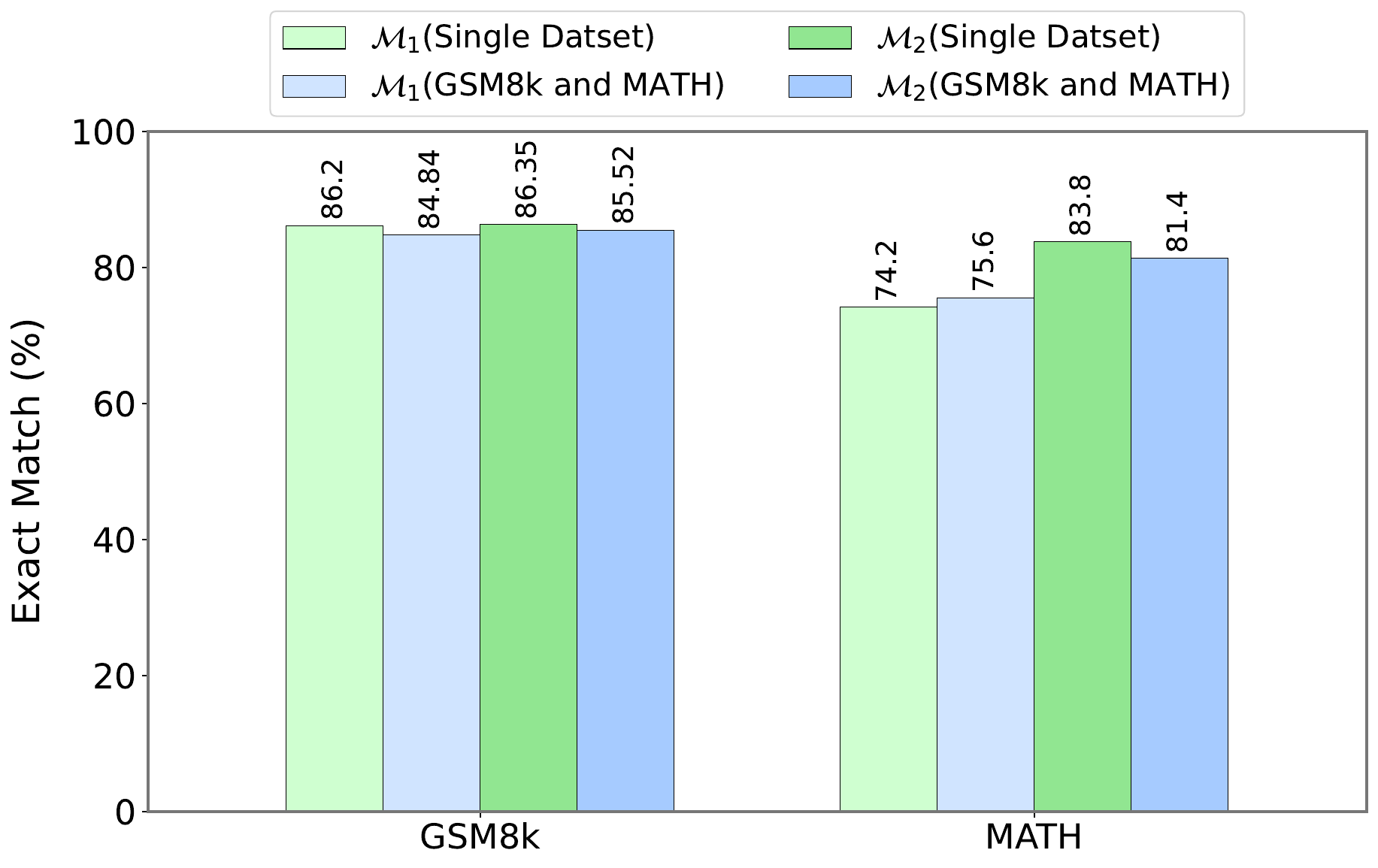}
    \caption{Comparison of plan-tuning accuracy on GSM8K and MATH when trained on each dataset individually \textit{vs.} their combined corpus. $\mathcal{M}_i$ (Single Dataset) indicates that the respective method is trained on a given dataset (green), while other indicates that the respective method is trained on both datasets (blue).}
    \label{fig:analysis_1}
\end{figure}

\subsection{Analysis}

\paragraph{Synthesis and Distribution-shift Considerations}
Together, these results indicate that embedding explicit natural planning into smaller LLMs—first via supervised trajectories and then through an RL-based policy refinement—yields substantial improvements in both in-domain accuracy and out-of-domain robustness. The significant improvements on AIME highlight how enforcing intermediate correctness reduces arithmetic drift, while consistent gains across two model families indicate the generality of our \textsc{plan-tuning} post-training approach in bridging distributional gaps between training and evaluation domains. 

\paragraph{Longer Reasoning Chains on OlympiadBench}
The reason behind the higher performance of \textsc{plan-tuned} models is that SFT provides clear templates for breaking down multi‐step proofs, giving the model a reliable blueprint for structured decomposition. Plan-GRPO builds on this by rewarding diverse, high-quality plan, encouraging the model to flexibly combine reasoning fragments when it encounters novel or unexpected subgoals.

\begin{figure*}
    \centering
    \includegraphics[width=0.8\linewidth]{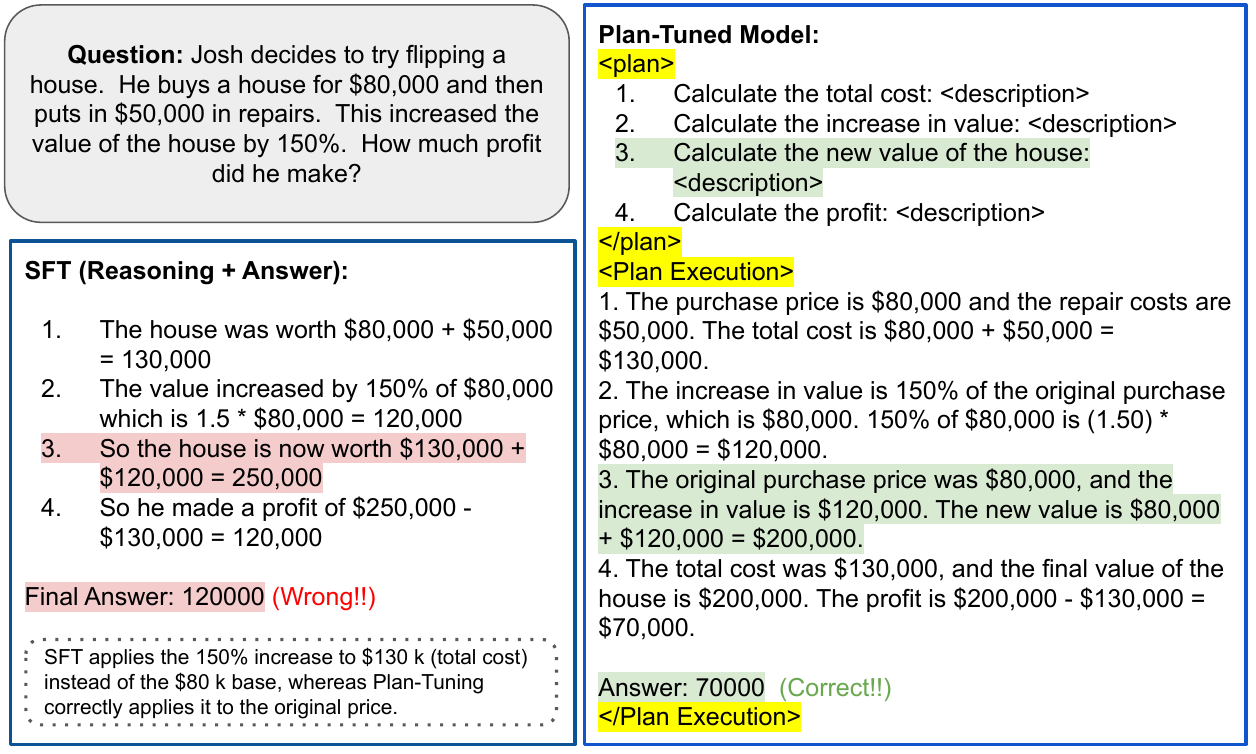}
    \caption{Qualitative Example: Comparison of SFT vs. \textsc{Plan‐Tuning} on a sample problem from GSM8k.}
    \label{fig:qualitative}
\end{figure*}

% \paragraph{Model Overfitting vs. Generalization}
% SFT can become tied to the specific plan patterns in the synthetic corpus, leading to lower performance under distribution shifts. In contrast, Plan-GRPO’s policy-gradient updates balance exploitation of known strategies with exploration of new ones, showing more robust generalization on out-of-domain benchmarks.

\paragraph{Variance in Plan Quality}
For short, arithmetic-focused GSM8K tasks, both methods quickly converge to high accuracy using straightforward reasoning chains. On more complex olympiad-level tasks—like geometry or combinatorics subcases—the RL objective in Plan-GRPO improves the impact of diverse, intermediate-valid plans, enabling the discovery of novel solution paths.

\paragraph{Mixing GSM8k and MATH during \textsc{plan-tuning}}

Figure \ref{fig:analysis_1} compares \textsc{plan-tuning} on GSM8K and MATH when training on each dataset separately\textit{vs.} training on their combination. In both the joint plan-and-answer ($\mathcal{M}_1$) and plan-only (($\mathcal{M}_2$)) variants, tuning on a single dataset yields the best accuracy on its own benchmark, 86.2\% on GSM8K and 83.8\% on MATH—whereas mixing the two corpora in a sample batch drops GSM8K performance by $\sim2\%$ and fails to improve MATH beyond its single-dataset result. This consistent degradation (except slight improvement in the $\mathcal{M}_1$(GSM8k and MATH)) indicates that \textsc{plan-tuning} relies on dataset-specific patterns of problem structure and reasoning style; when these patterns become heterogeneous, the model struggles to internalize a coherent planning policy, so domain-focused tuning can be more effective. The slight improvement in the $\mathcal{M}_1$(GSM8k and MATH) ($75.6\%$) compared to only $\mathcal{M}_1$ (MATH) ($74.2\%$) may seem to contradict our broader claim based on other results in Figure \ref{fig:analysis_1}. However, in $\mathcal{M}_1$, the model jointly learns plan, execution, and answer generation. This may allow the model to benefit slightly from additional GSM8K examples, which, although simpler, still provide useful structure for learning generalizable output formatting.

\begin{table}
\centering
\begin{tabular}{lrr}
\toprule
Method                & GSM8k & AIME 2024 \\
\midrule
$\mathcal{M}$(single-stage)      & 83.32 & 22.72     \\
$\mathcal{M}$(two-stage)         & \textbf{87.11} & \textbf{32.05}     \\
\bottomrule
\end{tabular}
\caption{Performance comparison of model trained with data created using single-stage \textit{vs.} two-stage filtering methods on GSM8k and AIME 2024.}
\label{tab:method_performance}
\end{table}

\paragraph{Importance of two-stage filtering for data synthesis} We conduct an ablation study in which we replace the two-stage filtering with using only the Plan Verifier (Figure \ref{fig:method_1}) threshold to select better plans for training. We performed experiments using Qwen3-8B on GSM8K for $\mathcal{M}_1$, where the training data was filtered only based on the Plan Verifier. We evaluate it on GSM8k and one OOD dataset, AIME 2024. Here, $\mathcal{M}$(single-stage) indicates the model trained with data created using single-stage filtering, and $\mathcal{M}$(two-stage) indicates the model trained with data created using two-stage filtering. The results from Table \ref{tab:method_performance} showcase that two-stage filtering leads to better improvements: a $\sim4\%$ improvement on GSM8K and $\sim9\%$ on AIME 2024. These findings support that two-stage filtering plays a critical role in improving the performance. This ablation shows the importance of combining plan quality with answer correctness for synthesizing good training data.

\paragraph{Qualitative Analysis} We provide an example of how \textsc{Plan-tuned} models improve reasoning and problem-solving capabilities over SFT in Figure \ref{fig:qualitative}. In this example, we demonstrate how planning trajectories steer the model toward a structured solution. The vanilla SFT model misapplies the 150\% increase to the combined cost, yielding an incorrect profit of \$120,000. In contrast, the plan‐tuned model explicitly decomposes the task into four subtasks—total cost, value increase, new house value, and profit—and arrives at the correct answer of \$70,000. The SFT pipeline collapses subtasks and propagates an early error, whereas our \textsc{Plan‐Tuning} framework enforces step-by-step reasoning aligned with human planning. This case highlights the importance of decomposing complex problems into clear intermediate goals to improve both accuracy and interoperability.

% in App. \ref{app:quality_example}.

\section{Conclusions}

We introduce \textsc{PLAN-TUNING}, a novel post-training method that incorporates synthetic natural planning trajectories into smaller LLMs’ parameters, rather than relying solely on inference-time prompts. We develop two complementary post-training strategies—SFT to imitate high-quality plan decompositions and GRPO to reinforce plan quality alongside answer correctness—thereby teaching models both how to plan and how to execute. Across two in-domain datasets (GSM8K, MATH), \textsc{PLAN-TUNED} models achieve an average $\sim7\%$ accuracy boost over strong baselines; on out-of-domain benchmarks (OlympiadBench, AIME), they significantly improved the performance. Through detailed analyses, we show (i) that a good plan relies on dataset-specific consistency—mixing heterogeneous sources degrades performance, and (ii) that plan-tuning substantially mitigates the token-repetition issue.

\section*{Limitations}

Our approach relies on a large base LLM to generate and verify planning trajectories; errors or biases in that upstream model can propagate into training data and limit downstream gains. Generating, filtering, and executing multiple candidate plans per example incurs non-trivial computational cost and implementation complexity, which may hinder large-scale or real-time applications. All experiments focus on mathematical reasoning; it remains to be validated whether \textsc{PLAN-TUNING} generalizes to other problem types (e.g., commonsense, code synthesis) without substantial adaptation. Key thresholds (e.g., plan-quality cutoff, reward weighting) require manual tuning and may not transfer directly across datasets or languages. For GRPO, we used only LLM-based verification for planning reward, we will certainly explore more objective-function evaluations such as ROSCOE \cite{golovnevaroscoe} as future work.

\section*{Ethics Statement}

The use of proprietary LLMs such as GPT-4, Gemini, and Claude-3 in this study adheres to their policies of usage. We have used AI assistants (Grammarly and Gemini) to address the grammatical errors and rephrase the sentences.

\bibliography{custom}

\begin{thebibliography}{37}
\providecommand{\natexlab}[1]{#1}

\bibitem[{Brown et~al.(2024)Brown, Juravsky, Ehrlich, Clark, Le, R{\'e}, and Mirhoseini}]{brown2024large}
Bradley Brown, Jordan Juravsky, Ryan Ehrlich, Ronald Clark, Quoc~V Le, Christopher R{\'e}, and Azalia Mirhoseini. 2024.
\newblock Large language monkeys: Scaling inference compute with repeated sampling.
\newblock \emph{arXiv preprint arXiv:2407.21787}.

\bibitem[{Chen et~al.(2024)Chen, Liu, Wang, Zhang, Liu, Lin, Chen, and Zhao}]{chen-etal-2024-agent}
Zehui Chen, Kuikun Liu, Qiuchen Wang, Wenwei Zhang, Jiangning Liu, Dahua Lin, Kai Chen, and Feng Zhao. 2024.
\newblock \href {https://doi.org/10.18653/v1/2024.findings-acl.557} {Agent-{FLAN}: Designing data and methods of effective agent tuning for large language models}.
\newblock In \emph{Findings of the Association for Computational Linguistics: ACL 2024}, pages 9354--9366, Bangkok, Thailand. Association for Computational Linguistics.

\bibitem[{Cobbe et~al.(2021)Cobbe, Kosaraju, Bavarian, Chen, Jun, Kaiser, Plappert, Tworek, Hilton, Nakano et~al.}]{cobbe2021training}
Karl Cobbe, Vineet Kosaraju, Mohammad Bavarian, Mark Chen, Heewoo Jun, Lukasz Kaiser, Matthias Plappert, Jerry Tworek, Jacob Hilton, Reiichiro Nakano, and 1 others. 2021.
\newblock Training verifiers to solve math word problems, 2021.
\newblock \emph{URL https://arxiv. org/abs/2110.14168}, 9.

\bibitem[{Golovneva et~al.(2023)Golovneva, Chen, Poff, Corredor, Zettlemoyer, Fazel-Zarandi, and Celikyilmaz}]{golovnevaroscoe}
Olga Golovneva, Moya~Peng Chen, Spencer Poff, Martin Corredor, Luke Zettlemoyer, Maryam Fazel-Zarandi, and Asli Celikyilmaz. 2023.
\newblock Roscoe: A suite of metrics for scoring step-by-step reasoning.
\newblock In \emph{The Eleventh International Conference on Learning Representations}.

\bibitem[{He et~al.(2024)He, Luo, Bai, Hu, Thai, Shen, Hu, Han, Huang, Zhang et~al.}]{he2024olympiadbench}
Chaoqun He, Renjie Luo, Yuzhuo Bai, Shengding Hu, Zhen Thai, Junhao Shen, Jinyi Hu, Xu~Han, Yujie Huang, Yuxiang Zhang, and 1 others. 2024.
\newblock Olympiadbench: A challenging benchmark for promoting agi with olympiad-level bilingual multimodal scientific problems.
\newblock In \emph{Proceedings of the 62nd Annual Meeting of the Association for Computational Linguistics (Volume 1: Long Papers)}, pages 3828--3850.

\bibitem[{Hendrycks et~al.(2021)Hendrycks, Burns, Kadavath, Arora, Basart, Tang, Song, and Steinhardt}]{hendrycks2021measuring}
Dan Hendrycks, Collin Burns, Saurav Kadavath, Akul Arora, Steven Basart, Eric Tang, Dawn Song, and Jacob Steinhardt. 2021.
\newblock \href {https://openreview.net/forum?id=7Bywt2mQsCe} {Measuring mathematical problem solving with the {MATH} dataset}.
\newblock In \emph{Thirty-fifth Conference on Neural Information Processing Systems Datasets and Benchmarks Track (Round 2)}.

\bibitem[{Jiao et~al.(2024)Jiao, Qin, Liu, Chen, and Joty}]{jiao-etal-2024-learning}
Fangkai Jiao, Chengwei Qin, Zhengyuan Liu, Nancy~F. Chen, and Shafiq Joty. 2024.
\newblock \href {https://doi.org/10.18653/v1/2024.emnlp-main.20} {Learning planning-based reasoning by trajectories collection and process reward synthesizing}.
\newblock In \emph{Proceedings of the 2024 Conference on Empirical Methods in Natural Language Processing}, pages 334--350, Miami, Florida, USA. Association for Computational Linguistics.

\bibitem[{Kojima et~al.(2022)Kojima, Gu, Reid, Matsuo, and Iwasawa}]{kojima2022large}
Takeshi Kojima, Shixiang~Shane Gu, Machel Reid, Yutaka Matsuo, and Yusuke Iwasawa. 2022.
\newblock Large language models are zero-shot reasoners.
\newblock \emph{Advances in neural information processing systems}, 35:22199--22213.

\bibitem[{Kuznia et~al.(2022)Kuznia, Mishra, Parmar, and Baral}]{kuznia2022less}
Kirby Kuznia, Swaroop Mishra, Mihir Parmar, and Chitta Baral. 2022.
\newblock Less is more: Summary of long instructions is better for program synthesis.
\newblock In \emph{Proceedings of the 2022 Conference on Empirical Methods in Natural Language Processing}, pages 4532--4552.

\bibitem[{Liu et~al.(2024)Liu, Feng, Xue, Wang, Wu, Lu, Zhao, Deng, Zhang, Ruan et~al.}]{liu2024deepseek}
Aixin Liu, Bei Feng, Bing Xue, Bingxuan Wang, Bochao Wu, Chengda Lu, Chenggang Zhao, Chengqi Deng, Chenyu Zhang, Chong Ruan, and 1 others. 2024.
\newblock Deepseek-v3 technical report.
\newblock \emph{arXiv preprint arXiv:2412.19437}.

\bibitem[{Liu et~al.(2023)Liu, Yuan, Fu, Jiang, Hayashi, and Neubig}]{liu2023pre}
Pengfei Liu, Weizhe Yuan, Jinlan Fu, Zhengbao Jiang, Hiroaki Hayashi, and Graham Neubig. 2023.
\newblock Pre-train, prompt, and predict: A systematic survey of prompting methods in natural language processing.
\newblock \emph{ACM Computing Surveys}, 55(9):1--35.

\bibitem[{Ouyang et~al.(2022)Ouyang, Wu, Jiang, Almeida, Wainwright, Mishkin, Zhang, Agarwal, Slama, Ray et~al.}]{ouyang2022training}
Long Ouyang, Jeffrey Wu, Xu~Jiang, Diogo Almeida, Carroll Wainwright, Pamela Mishkin, Chong Zhang, Sandhini Agarwal, Katarina Slama, Alex Ray, and 1 others. 2022.
\newblock Training language models to follow instructions with human feedback.
\newblock \emph{Advances in neural information processing systems}, 35:27730--27744.

\bibitem[{Parmar et~al.(2025)Parmar, Liu, Goyal, Chen, Le, Mishra, Mobahi, Gu, Wang, Nakhost et~al.}]{parmar2025plangen}
Mihir Parmar, Xin Liu, Palash Goyal, Yanfei Chen, Long Le, Swaroop Mishra, Hossein Mobahi, Jindong Gu, Zifeng Wang, Hootan Nakhost, and 1 others. 2025.
\newblock Plangen: A multi-agent framework for generating planning and reasoning trajectories for complex problem solving.
\newblock \emph{arXiv preprint arXiv:2502.16111}.

\bibitem[{Patel et~al.(2022)Patel, Mishra, Parmar, and Baral}]{patel2022question}
Pruthvi Patel, Swaroop Mishra, Mihir Parmar, and Chitta Baral. 2022.
\newblock Is a question decomposition unit all we need?
\newblock In \emph{Proceedings of the 2022 Conference on Empirical Methods in Natural Language Processing}, pages 4553--4569.

\bibitem[{Rafailov et~al.(2023)Rafailov, Sharma, Mitchell, Manning, Ermon, and Finn}]{rafailov2023direct}
Rafael Rafailov, Archit Sharma, Eric Mitchell, Christopher~D Manning, Stefano Ermon, and Chelsea Finn. 2023.
\newblock \href {https://openreview.net/forum?id=HPuSIXJaa9} {Direct preference optimization: Your language model is secretly a reward model}.
\newblock In \emph{Thirty-seventh Conference on Neural Information Processing Systems}.

\bibitem[{Rein et~al.(2024)Rein, Hou, Stickland, Petty, Pang, Dirani, Michael, and Bowman}]{rein2024gpqa}
David Rein, Betty~Li Hou, Asa~Cooper Stickland, Jackson Petty, Richard~Yuanzhe Pang, Julien Dirani, Julian Michael, and Samuel~R Bowman. 2024.
\newblock Gpqa: A graduate-level google-proof q\&a benchmark.
\newblock In \emph{First Conference on Language Modeling}.

\bibitem[{Saeidi et~al.(2024)Saeidi, Verma, RRV, and Baral}]{saeidi2024triple}
Amir Saeidi, Shivanshu Verma, Aswin RRV, and Chitta Baral. 2024.
\newblock Triple preference optimization: Achieving better alignment with less data in a single step optimization.
\newblock \emph{arXiv preprint arXiv:2405.16681}.

\bibitem[{Schulman et~al.(2017)Schulman, Wolski, Dhariwal, Radford, and Klimov}]{schulman2017proximal}
John Schulman, Filip Wolski, Prafulla Dhariwal, Alec Radford, and Oleg Klimov. 2017.
\newblock Proximal policy optimization algorithms.
\newblock \emph{arXiv preprint arXiv:1707.06347}.

\bibitem[{Shao et~al.(2024)Shao, Wang, Zhu, Xu, Song, Bi, Zhang, Zhang, Li, Wu et~al.}]{shao2024deepseekmath}
Zhihong Shao, Peiyi Wang, Qihao Zhu, Runxin Xu, Junxiao Song, Xiao Bi, Haowei Zhang, Mingchuan Zhang, YK~Li, Y~Wu, and 1 others. 2024.
\newblock Deepseekmath: Pushing the limits of mathematical reasoning in open language models.
\newblock \emph{arXiv preprint arXiv:2402.03300}.

\bibitem[{Snell et~al.(2024)Snell, Lee, Xu, and Kumar}]{snell2024scaling}
Charlie Snell, Jaehoon Lee, Kelvin Xu, and Aviral Kumar. 2024.
\newblock Scaling llm test-time compute optimally can be more effective than scaling model parameters.
\newblock \emph{arXiv preprint arXiv:2408.03314}.

\bibitem[{Song et~al.(2024{\natexlab{a}})Song, Xiong, Zhao, Zhu, Wu, Wang, Li, Peng, and Li}]{song-etal-2024-agentbank}
Yifan Song, Weimin Xiong, Xiutian Zhao, Dawei Zhu, Wenhao Wu, Ke~Wang, Cheng Li, Wei Peng, and Sujian Li. 2024{\natexlab{a}}.
\newblock \href {https://doi.org/10.18653/v1/2024.findings-emnlp.116} {{A}gent{B}ank: Towards generalized {LLM} agents via fine-tuning on 50000+ interaction trajectories}.
\newblock In \emph{Findings of the Association for Computational Linguistics: EMNLP 2024}, pages 2124--2141, Miami, Florida, USA. Association for Computational Linguistics.

\bibitem[{Song et~al.(2024{\natexlab{b}})Song, Yin, Yue, Huang, Li, and Lin}]{song-etal-2024-trial}
Yifan Song, Da~Yin, Xiang Yue, Jie Huang, Sujian Li, and Bill~Yuchen Lin. 2024{\natexlab{b}}.
\newblock \href {https://doi.org/10.18653/v1/2024.acl-long.409} {Trial and error: Exploration-based trajectory optimization of {LLM} agents}.
\newblock In \emph{Proceedings of the 62nd Annual Meeting of the Association for Computational Linguistics (Volume 1: Long Papers)}, pages 7584--7600, Bangkok, Thailand. Association for Computational Linguistics.

\bibitem[{Sun et~al.(2025)Sun, Min, Chen, Zhao, Liu, Wang, Fang, and Wen}]{sun2025challenging}
Haoxiang Sun, Yingqian Min, Zhipeng Chen, Wayne~Xin Zhao, Zheng Liu, Zhongyuan Wang, Lei Fang, and Ji-Rong Wen. 2025.
\newblock Challenging the boundaries of reasoning: An olympiad-level math benchmark for large language models.
\newblock \emph{arXiv preprint arXiv:2503.21380}.

\bibitem[{Team et~al.(2023)Team, Anil, Borgeaud, Alayrac, Yu, Soricut, Schalkwyk, Dai, Hauth, Millican et~al.}]{team2023gemini}
Gemini Team, Rohan Anil, Sebastian Borgeaud, Jean-Baptiste Alayrac, Jiahui Yu, Radu Soricut, Johan Schalkwyk, Andrew~M Dai, Anja Hauth, Katie Millican, and 1 others. 2023.
\newblock Gemini: a family of highly capable multimodal models.
\newblock \emph{arXiv preprint arXiv:2312.11805}.

\bibitem[{Team et~al.(2025)Team, Kamath, Ferret, Pathak, Vieillard, Merhej, Perrin, Matejovicova, Ram{\'e}, Rivi{\`e}re et~al.}]{team2025gemma}
Gemma Team, Aishwarya Kamath, Johan Ferret, Shreya Pathak, Nino Vieillard, Ramona Merhej, Sarah Perrin, Tatiana Matejovicova, Alexandre Ram{\'e}, Morgane Rivi{\`e}re, and 1 others. 2025.
\newblock Gemma 3 technical report.
\newblock \emph{arXiv preprint arXiv:2503.19786}.

\bibitem[{Wang et~al.(2024{\natexlab{a}})Wang, Cassano, Wu, Bai, Song, Nath, Han, Hendryx, Yue, and Zhang}]{wang2planning}
Evan~Z Wang, Federico Cassano, Catherine Wu, Yunfeng Bai, William Song, Vaskar Nath, Ziwen Han, Sean~M Hendryx, Summer Yue, and Hugh Zhang. 2024{\natexlab{a}}.
\newblock Planning in natural language improves llm search for code generation.
\newblock In \emph{The First Workshop on System-2 Reasoning at Scale, NeurIPS'24}.

\bibitem[{Wang et~al.(2023)Wang, Xu, Lan, Hu, Lan, Lee, and Lim}]{wang2023plan}
Lei Wang, Wanyu Xu, Yihuai Lan, Zhiqiang Hu, Yunshi Lan, Roy Ka-Wei Lee, and Ee-Peng Lim. 2023.
\newblock Plan-and-solve prompting: Improving zero-shot chain-of-thought reasoning by large language models.
\newblock In \emph{Proceedings of the 61st Annual Meeting of the Association for Computational Linguistics (Volume 1: Long Papers)}, pages 2609--2634.

\bibitem[{Wang et~al.(2024{\natexlab{b}})Wang, Li, Han, Zhang, and Baldwin}]{wang2024learning}
Renxi Wang, Haonan Li, Xudong Han, Yixuan Zhang, and Timothy Baldwin. 2024{\natexlab{b}}.
\newblock Learning from failure: Integrating negative examples when fine-tuning large language models as agents.
\newblock \emph{arXiv preprint arXiv:2402.11651}.

\bibitem[{Wang et~al.(2022)Wang, Wei, Schuurmans, Le, Chi, Narang, Chowdhery, and Zhou}]{wang2022self}
Xuezhi Wang, Jason Wei, Dale Schuurmans, Quoc Le, Ed~Chi, Sharan Narang, Aakanksha Chowdhery, and Denny Zhou. 2022.
\newblock Self-consistency improves chain of thought reasoning in language models.
\newblock \emph{arXiv preprint arXiv:2203.11171}.

\bibitem[{Wei et~al.(2022)Wei, Wang, Schuurmans, Bosma, Xia, Chi, Le, Zhou et~al.}]{wei2022chain}
Jason Wei, Xuezhi Wang, Dale Schuurmans, Maarten Bosma, Fei Xia, Ed~Chi, Quoc~V Le, Denny Zhou, and 1 others. 2022.
\newblock Chain-of-thought prompting elicits reasoning in large language models.
\newblock \emph{Advances in neural information processing systems}, 35:24824--24837.

\bibitem[{Wu et~al.(2024)Wu, Sun, Li, Welleck, and Yang}]{wu2024empirical}
Yangzhen Wu, Zhiqing Sun, Shanda Li, Sean Welleck, and Yiming Yang. 2024.
\newblock An empirical analysis of compute-optimal inference for problem-solving with language models.
\newblock \emph{arXiv preprint arXiv:2408.00724}.

\bibitem[{Yang et~al.(2024)Yang, Yang, Zhang, Hui, Zheng, Yu, Li, Liu, Huang, Wei, Lin, Yang, Tu, Zhang, Yang, Yang, Zhou, Lin, Dang, Lu, Bao, Yang, Yu, Li, Xue, Zhang, Zhu, Men, Lin, Li, Xia, Ren, Ren, Fan, Su, Zhang, Wan, Liu, Cui, Zhang, and Qiu}]{qwen2.5}
An~Yang, Baosong Yang, Beichen Zhang, Binyuan Hui, Bo~Zheng, Bowen Yu, Chengyuan Li, Dayiheng Liu, Fei Huang, Haoran Wei, Huan Lin, Jian Yang, Jianhong Tu, Jianwei Zhang, Jianxin Yang, Jiaxi Yang, Jingren Zhou, Junyang Lin, Kai Dang, and 22 others. 2024.
\newblock Qwen2.5 technical report.
\newblock \emph{arXiv preprint arXiv:2412.15115}.

\bibitem[{Zhang et~al.(2024{\natexlab{a}})Zhang, Huang, Zhou, Li, and Ouyang}]{zhang2024accessing}
Di~Zhang, Xiaoshui Huang, Dongzhan Zhou, Yuqiang Li, and Wanli Ouyang. 2024{\natexlab{a}}.
\newblock Accessing gpt-4 level mathematical olympiad solutions via monte carlo tree self-refine with llama-3 8b.
\newblock \emph{arXiv preprint arXiv:2406.07394}.

\bibitem[{Zhang et~al.(2024{\natexlab{b}})Zhang, Lan, Rithesh, Liu, Yao, Tan, Hoang, Yang, Feng, Liu et~al.}]{zhang2024agent}
Jianguo Zhang, Tian Lan, RN~Rithesh, Zhiwei Liu, Weiran Yao, Juntao Tan, Thai~Quoc Hoang, Liangwei Yang, Yihao Feng, Zuxin Liu, and 1 others. 2024{\natexlab{b}}.
\newblock The agent ohana: Designing unified data and training pipeline for effective agent learning.
\newblock In \emph{ICLR 2024 Workshop on Large Language Model (LLM) Agents}.

\bibitem[{Zheng et~al.(2024)Zheng, Mishra, Zhang, Chen, Chen, Nova, Hou, Cheng, Le, Chi et~al.}]{zheng2024natural}
Huaixiu~Steven Zheng, Swaroop Mishra, Hugh Zhang, Xinyun Chen, Minmin Chen, Azade Nova, Le~Hou, Heng-Tze Cheng, Quoc~V Le, Ed~H Chi, and 1 others. 2024.
\newblock Natural plan: Benchmarking llms on natural language planning.
\newblock \emph{arXiv preprint arXiv:2406.04520}.

\bibitem[{Zhong et~al.(2024)Zhong, Liu, Pan, Zhang, Zhou, Liang, Wu, Lyu, Shu, Yu et~al.}]{zhong2024evaluation}
Tianyang Zhong, Zhengliang Liu, Yi~Pan, Yutong Zhang, Yifan Zhou, Shizhe Liang, Zihao Wu, Yanjun Lyu, Peng Shu, Xiaowei Yu, and 1 others. 2024.
\newblock Evaluation of openai o1: Opportunities and challenges of agi.
\newblock \emph{arXiv preprint arXiv:2409.18486}.

\bibitem[{Zhou et~al.(2023)Zhou, Sch{\"a}rli, Hou, Wei, Scales, Wang, Schuurmans, Cui, Bousquet, Le et~al.}]{zhouleast}
Denny Zhou, Nathanael Sch{\"a}rli, Le~Hou, Jason Wei, Nathan Scales, Xuezhi Wang, Dale Schuurmans, Claire Cui, Olivier Bousquet, Quoc~V Le, and 1 others. 2023.
\newblock Least-to-most prompting enables complex reasoning in large language models.
\newblock In \emph{The Eleventh International Conference on Learning Representations}.

\end{thebibliography}

\clearpage

\appendix

\section{Example Prompts and Responses}
\label{app:examples_planning}

These prompts exemplify everyday reasoning tasks that require multi-step planning and decision-making, making them ideal for evaluating LLMs' ability to perform structured, goal-oriented reasoning. From organizing a trip or scheduling a meeting to planning a birthday party or solving a math problem, each scenario involves decomposing a high-level objective into coherent subtasks. The corresponding responses reflect how a well-structured plan can guide users through complex decisions—highlighting both procedural thinking and domain knowledge. Such prompts can be used as qualitative case studies to assess models' competence in natural planning, temporal sequencing, and commonsense reasoning.

\subsection*{How to plan a trip to Japan}

\begin{tcolorbox}[colframe=blue!50!black,title=Response, breakable]
\small
\begin{enumerate}[leftmargin=*]
    \item \textbf{Decide on Travel Dates} – Choose the season (e.g., cherry blossom in spring or autumn foliage) and trip duration.
    \item \textbf{Set a Budget} – Estimate total costs including flights, transport, lodging, food, and attractions.
    \item \textbf{Check Entry Requirements} – Look into visa policies, passport validity, and any travel restrictions.
    \item \textbf{Choose Destinations} – Select cities and regions (e.g., Tokyo, Kyoto, Osaka, Nara, Hokkaido, Okinawa) based on your interests.
    \item \textbf{Book Flights and Accommodation} – Use platforms like Google Flights, Booking.com, or Airbnb; book early in peak seasons.
    \item \textbf{Plan Transportation Within Japan} – Consider a Japan Rail Pass and IC cards (Suica, Pasmo) for local travel.
    \item \textbf{Create an Itinerary} – Map out a day-by-day plan including travel time, attractions, and rest periods.
    \item \textbf{Arrange Travel Essentials} – Pocket Wi-Fi/SIM card, power adapter (Type A/B), Japanese yen (cash is still widely used).
    \item \textbf{Learn Local Etiquette} – Study basic customs (e.g., removing shoes indoors, quiet on trains, no tipping).
    \item \textbf{Pack Appropriately} – Bring weather-appropriate clothing, walking shoes, and a small daypack for city exploration.
\end{enumerate}
\end{tcolorbox}

\subsection*{How to schedule a meeting with your PhD colleague}

\begin{tcolorbox}[colframe=blue!50!black,title=Response, breakable]
\small
\begin{enumerate}[leftmargin=*]
    \item \textbf{Define the Purpose} – Clarify the topic (e.g., paper discussion, experiment planning).
    \item \textbf{Propose a Timeframe} – Suggest a few potential date/time slots that work for you.
    \item \textbf{Use Scheduling Tools} – Use Doodle, Google Calendar, or Outlook to find mutual availability.
    \item \textbf{Check Their Availability} – Send a quick message (email/Slack/WhatsApp) asking for their availability.
    \item \textbf{Suggest a Duration} – Be clear about how long you expect the meeting to take (30 mins, 1 hour, etc.).
    \item \textbf{Decide the Format} – In-person or virtual (Zoom, Teams, Google Meet).
    \item \textbf{Send Calendar Invite} – Include topic, agenda (if needed), and location or meeting link.
    \item \textbf{Follow Up Politely} – If they don’t respond in a day or two, send a short, polite reminder.
    \item \textbf{Prepare Materials} – If needed, send any relevant documents or reading material beforehand.
    \item \textbf{Confirm on the Day} – Send a quick message confirming the meeting a few hours before, especially for virtual calls.
\end{enumerate}
\end{tcolorbox}

\subsection*{How to organize a party for the birthday}

\begin{tcolorbox}[colframe=blue!50!black,title=Response, breakable]
\small
\begin{enumerate}[leftmargin=*]
    \item \textbf{Set a Budget} – Decide how much you want to spend overall.
    \item \textbf{Choose a Date \& Time} – Pick a convenient day for the birthday person and guests.
    \item \textbf{Pick a Venue} – At home, a restaurant, park, or rented space—based on budget and guest size.
    \item \textbf{Decide a Theme (optional)} – Themes like retro, beach, movie night, or favorite colors add fun.
    \item \textbf{Make a Guest List} – Know how many people you’re inviting to plan food and space.
    \item \textbf{Send Invitations} – Use digital invites (WhatsApp, email, Evite) or printed cards.
    \item \textbf{Plan Food \& Drinks} – Homemade, catered, or ordered—match it to the guest profile and theme.
    \item \textbf{Arrange Decorations \& Cake} – Buy or DIY decorations and order a cake in advance.
    \item \textbf{Plan Activities or Entertainment} – Music, games, speeches, or photo booth to keep guests engaged.
    \item \textbf{Prepare a Backup Plan} – Especially for outdoor events, have a Plan B in case of weather or other issues.
\end{enumerate}
\end{tcolorbox}

\subsection*{How to solve a given mathematical reasoning problem}

\begin{tcolorbox}[colframe=blue!50!black,title=Response, breakable]
\small
\begin{enumerate}[leftmargin=*]
    \item \textbf{Understand the Problem} – Read carefully and identify what is being asked.
    \item \textbf{Identify Key Information} – Extract numbers, relationships, and conditions from the problem.
    \item \textbf{Break Down the Problem} – Divide the complex problem into smaller, manageable parts.
    \item \textbf{Visualize or Represent} – Use diagrams, equations, or tables if helpful.
    \item \textbf{Recall Relevant Concepts} – Think of formulas, theorems, or reasoning strategies that apply.
    \item \textbf{Formulate a Plan} – Decide on the method (e.g., algebra, logic, geometry) to approach the problem.
    \item \textbf{Execute Step-by-Step} – Carry out your plan methodically, solving subparts if needed.
    \item \textbf{Check Units and Logic} – Make sure the answer is in the correct form and logically consistent.
    \item \textbf{Verify Your Answer} – Plug it back into the original problem to confirm it satisfies all conditions.
    \item \textbf{Reflect on the Method} – Consider if there’s a more efficient or alternative solution strategy.
\end{enumerate}
\end{tcolorbox}

\section{Data Synthesis Prompts}
\label{app:data_prompt}

These prompts form a structured framework for evaluating and improving mathematical reasoning in large language models. The \textit{Plan Generation Prompt} encourages models to decompose complex math problems into step-by-step solution strategies, fostering procedural thinking. The \textit{Constraints Generation Prompt} identifies key logical and mathematical conditions that must be satisfied by any valid solution plan, serving as a verification checklist. Finally, the \textit{Plan Verification Prompt} introduces a rigorous reward-based scoring scheme, allowing evaluators to assign interpretable, constraint-aware scores to the quality of generated plans. This framework promotes transparency, robustness, and fidelity in evaluating model reasoning capabilities.

\subsection*{Plan Generation Prompt}

\begin{tcolorbox}[colframe=blue!50!black,title=Prompt, breakable]
\small
Analyze the given maths question, and create a plan to solve it:

\begin{verbatim}
<question>
{question}
</question>
\end{verbatim}

Feel free to break down the problem in whatever way you think is most effective. Consider key concepts, formulas, relevant facts, or any logical approach that would help solve this. Your task is to only provide a plan and not solve it during this process.
\end{tcolorbox}

\subsection*{Constraints Generation Prompt}

\begin{tcolorbox}[colframe=blue!50!black,title=Prompt, breakable]
\small
You are an expert in identifying explicit and implicit constraints for verifying plans generated to solve complex maths problems. Your job is to generate those constraints for the following question, which can be helpful in verifying and evaluating the given plan.

\begin{verbatim}
<question>
{question}
</question>
\end{verbatim}

Make sure to identify all constraints in the question. Please output the constraints as a list. DO NOT include any other text in your response.
\end{tcolorbox}

\subsection*{Plan Verification Prompt}

\begin{tcolorbox}[colframe=blue!50!black,title=Prompt, breakable]
\small
Provide a reward score between -100 and 100 for the plan quality, using very strict standards. Do not give a full score above 95. Make sure the reward score is an integer.

\textbf{Input:}
\begin{verbatim}
{input}
\end{verbatim}

\textbf{Generated plan to evaluate:}
\begin{verbatim}
{response}
\end{verbatim}

Consider constraints below while evaluating:
\begin{verbatim}
{verification_prompt}
\end{verbatim}

Make sure to check all the constraints before giving the reward score.

Please provide a reward in the format below:

\begin{itemize}[leftmargin=*]
    \item Steps: [step-by-step reasoning for the reward score]
    \item Score: [Strictly provide the reward score as an integer between -100 and 100]
\end{itemize}

\end{tcolorbox}

\section{Prompt for Planning Reward}
\label{app:planning_reward}

The box below defines the exact evaluation prompt we use to score planning quality in GRPO. It asks the model to compare a generated plan against a gold plan, provide a brief similarity analysis, and emit a single scalar score on a 0–1 scale. This score then serves directly as the planning reward during RL fine-tuning.

\begin{tcolorbox}[colframe=blue!50!black,title=Prompt, breakable]
\small

You are an expert evaluator of problem-solving plans. Compare the following two plans and rate their similarity on a scale from 0 to 1.

0.0 = Completely different plans with no shared approach or reasoning steps.  
0.25 = Minimal similarity with some overlapping concepts but fundamentally different approaches.  
0.5 = Moderate similarity with shared key ideas but significant differences in execution or reasoning.  
0.75 = High similarity with mostly aligned reasoning and steps, with minor differences.  
1.0 = Nearly identical plans that follow the same approach and reasoning steps.

\medskip
\textbf{Generated Plan:} \\
\{generated plan\}

\medskip
\textbf{Gold Plan:} \\
\{gold plan\}

\medskip
First, provide a brief analysis of the similarity, then output only a single float number between 0 and 1, representing the similarity score.

Please \textbf{STRICTLY} use the format below:

\begin{verbatim}
Analysis: [brief analysis]
Score: [float number between 0 and 1]
\end{verbatim}

\medskip
(Note: this score will be used as the planning reward in GRPO.)

\end{tcolorbox}

% \section{Qualitative Analysis}
% \label{app:quality_example}

% \begin{figure}[h]
%     \centering
%     \includegraphics[width=\linewidth]{figures/example_3.pdf}
%     \caption{Qualitative Example: Comparison of SFT vs. \textsc{Plan‐Tuning} on a sample problem from GSM8k.}
%     \label{fig:qualitative}
% \end{figure}

% In this example, we demonstrate how planning trajectories steer the model toward a structured solution. The vanilla SFT model misapplies the 150\% increase to the combined cost, yielding an incorrect profit of \$120,000. In contrast, the plan‐tuned model explicitly decomposes the task into four subtasks—total cost, value increase, new house value, and profit—and arrives at the correct answer of \$70,000. The SFT pipeline collapses subtasks and propagates an early error, whereas our \textsc{Plan‐Tuning} framework enforces step-by-step reasoning aligned with human planning. This case highlights the importance of decomposing complex problems into clear intermediate goals to improve both accuracy and interoperability.

% \section{Example Appendix}
% \label{sec:appendix}

% This is an appendix.

\end{document}